\documentclass[letterpaper, 10 pt, conference]{ieeeconf}  
\usepackage{amssymb}
\usepackage{amsmath} 
\usepackage{graphicx} 
\usepackage{wrapfig}
\usepackage{booktabs}
\usepackage{multirow}
\usepackage{cite}
\usepackage{adjustbox}   
\usepackage{makecell}  
\usepackage{url}  
\usepackage[hidelinks]{hyperref}
\usepackage[x11names]{xcolor}
\usepackage[caption=false,font=footnotesize]{subfig} 
\usepackage{booktabs}
\usepackage[font=footnotesize,labelfont=bf]{caption}
\IEEEoverridecommandlockouts                              

\newcommand{\methodname}{RoboSSM}

\newif\ifanon
\anontrue

\title{\methodname{}: Scalable In-context Imitation Learning via State-Space Models}

\author{
  Youngju Yoo$^{1,2}$ \quad
  Jiaheng Hu$^{1}$\quad
  Yifeng Zhu$^{1}$\quad
  Bo Liu$^{1,3}$\\
  Qiang Liu$^{1}$\quad
  Roberto Mart\'{\i}n-Mart\'{\i}n$^{1,4}$\quad
  Peter Stone$^{1,5}$ \\
  $^{1}$The University of Texas at Austin \quad
  $^{2}$KAIST \quad
  $^{3}$FAIR at Meta \quad
  $^{4}$Amazon \quad
  $^{5}$Sony AI \\
}

\begin{document}

\maketitle
\thispagestyle{empty}
\pagestyle{empty}

\begin{abstract}

In-context imitation learning (ICIL) enables robots to learn tasks from prompts consisting of just a handful of demonstrations. By eliminating the need for parameter updates at deployment time, this paradigm supports few-shot adaptation to novel tasks.
However, recent ICIL methods rely on Transformers, which have computational limitations and tend to underperform when handling longer prompts than those seen during training.
In this work, we introduce \methodname{}, a scalable recipe for in-context imitation learning based on state-space models (SSM).
Specifically, \methodname{} replaces Transformers with Longhorn -- a state-of-the-art SSM that provides linear-time inference and strong extrapolation capabilities, making it well-suited for long-context prompts.
Through diverse experiments on the LIBERO benchmark, we demonstrate the effectiveness of applying SSMs to ICIL, achieving improved generalization to both unseen and long-horizon tasks than Transformer-based ICIL methods by handling longer contexts at test-time.
These results show for the first time that SSMs are an efficient and scalable backbone for ICIL.
Our code is available at \href{https://github.com/youngjuY/RoboSSM}{\textcolor{RoyalBlue4}{https://github.com/youngjuY/RoboSSM}}.

\end{abstract}

\section{INTRODUCTION}

Imitation Learning (IL) is a powerful framework that enables robots to learn behaviors from demonstrations without explicit programming or reward design~\cite{il_argall,il_billard}.
While IL has achieved notable success in manipulation and navigation tasks, a key limitation of conventional imitation learning lies in its restricted adaptation capability, particularly when faced with new tasks.
Even with models trained on large multi-task datasets~\cite{octo_team,openvla_kim,openx_o,droid_khazatsky}, adapting to novel tasks still requires collecting a large amount of task-specific data and retraining, which can be computationally costly and often unstable~\cite{kim2025fine,hu2024flare}.
To address this challenge, In-Context Imitation Learning (ICIL) introduces a new paradigm, inspired by the success of large language models (LLMs)~\cite{gpt4_achiam,llama2_touvron,gemini2_comanici} in adapting to unseen language tasks through few-shot learning~\cite{fewshot_brown}. 
ICIL integrates the concept of prompting into imitation learning~\cite{fewshot_duan,fewshot_mandi,fewshot_valassakis,fewshot_xu,fewshot_brown,icrt_fu,lipvq_vuong,kat_palo}, allowing the model to infer and perform tasks based on a prompt composed of demonstrations, with no post-demonstration training.

Given that ICIL formulates imitation learning as a sequence modeling problem, recent ICIL approaches have naturally adopted Transformer-based models as their primary architecture~\cite{icrt_fu,kat_palo,lipvq_vuong}.
Although Transformers are the dominant architecture for sequence modeling~\cite{attention_vaswani}, their time complexity scales quadratically with sequence length, and they struggle to extrapolate beyond training lengths~\cite{transformer_length_zhou,flashattention_dao}.
For ICIL to handle long prompts efficiently at test time, it is essential to adopt alternatives to Transformers that enhance scalability with input length.

In this paper, we introduce \textbf{\methodname{}}, a scalable in-context learning framework that replaces Transformers with state-space models (SSMs).
Specifically, \methodname{} utilizes Longhorn~\cite{longhorn_liu}, a state-of-the-art SSM with linear inference time and strong extrapolation capability for long-context sequences.
Leveraging these properties, we show that \methodname{} can process substantially longer test-time prompts than previous Transformer-based ICIL methods.
We also investigate adapting Longhorn to ICIL via $\beta$-scaling ablations, which encourage the model to attend to demonstration prompts.

On the LIBERO~\cite{libero_liu} benchmark, we demonstrate that \methodname{} achieves both strong performance and efficient inference in long-context ICIL, addressing a key limitation of Transformer-based ICIL methods.
Specifically, \methodname{} uniquely benefits from longer prompts with additional in-context examples, maintaining high success rates on unseen tasks despite being trained with only a few demonstrations.
For instance, on the task \textit{pick up the plate and place it in the tray}, where the \textit{plate} object was unseen during training, \methodname{} achieves its highest performance when prompted with 32 demonstrations, despite being trained on only two.
Furthermore, we show that \methodname{} generalizes to unseen long-horizon tasks beyond the short-horizon training distribution, including multi-stage tasks and time-dilated scenarios created by repeating frames in the demonstrations.
Consequently, \methodname{} handles test-time demonstration prompts up to 16 times longer than those seen in training while maintaining linear inference time, thereby outperforming Transformer-based ICIL methods, which sharply degrade once the test prompt exceeds the training length.
These findings confirm that \methodname{} is the first approach to enable scalable in-context imitation learning by effectively leveraging long-range contextual information.
Our project video is available at \href{https://youtu.be/YR4m21zHDvM}{\textcolor{RoyalBlue4}{RoboSSM-video}}.

\begin{figure*}[t]
  \centering
  {\includegraphics[width=0.95\textwidth]{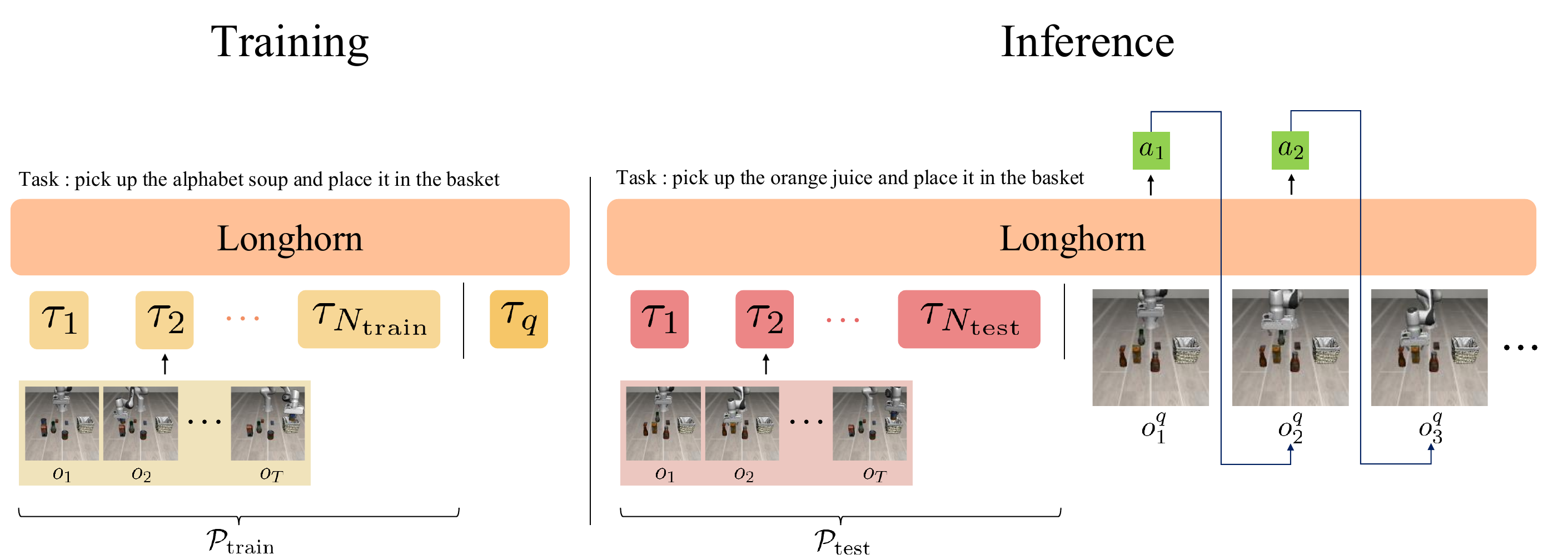}} 
  \caption{\textbf{Overview of \methodname{}}. \textbf{Training} (left): Longhorn receives $N_{\mathrm{train}}$ trajectories from $\mathcal{P}_{\mathrm{train}}$ and a query trajectory for the same task. \textbf{Inference} (right): Given $N_{\mathrm{test}}$ trajectories from $\mathcal{P}_{\mathrm{test}}$ containing unseen tasks, the model predicts actions and updates the environment iteratively from the initial observation embedding.}
  \label{fig:figure1}
\end{figure*}

\section{Related Work}

In this section, we provide an overview of prior ICIL methods and state-space models (SSMs), along with their recent applications to robotics.

\subsection{In-Context Imitation Learning}
Imitation learning~\cite{il_argall,il_billard,il_sun,il_wang} has long been a foundation for imparting skills to robots by learning from demonstration data.
Standard behavior cloning approaches~\cite{bc_bain,bc_torabi} typically train a separate policy for each task or rely on large multi-task datasets to acquire broader skills.
While multi-task imitation learning~\cite{octo_team,openvla_kim} can handle diverse tasks, these methods still struggle to perform completely unseen tasks without additional data collection for fine-tuning.

Inspired by the in-context learning paradigm in large language models (LLMs)~\cite{fewshot_brown}, recent imitation learning methods aim to eliminate parameter updates at test time, instead prompting a multi-task policy with a few demonstrations of unseen tasks.
Keypoint Action Tokens~\cite{kat_palo} introduce an ICIL framework that converts the visual observations and actions into tokens, which are fed into a pre-trained large language model.
ICRT~\cite{icrt_fu} performs in-context learning using a causal Transformer that predicts actions with next-token prediction, conditioned on a prompt consisting of a sequence of encoded teleoperated demonstrations.
LipVQ-VAE~\cite{lipvq_vuong} is an action tokenizer that uses vector quantization to address the lack of temporal smoothness in existing tokenizers and enable ICIL.

Although ICIL methods have been extensively studied and have achieved significant progress, they are typically trained and evaluated on test prompts that closely match the training prompt distribution in terms of length.
For instance, ICRT learns from inputs containing five demonstrations and masks the first random $k$ of them as the prompt, then at inference time is evaluated with three demonstration prompts.
LipVQ-VAE is trained and evaluated using only a single full demonstration as the prompt.
In contrast, \methodname{} explicitly aims to handle prompts that significantly deviate from the training distribution, such as those containing a larger number of demonstrations or long-horizon tasks.

\subsection{State-Space Models}

State-space models (SSMs) have emerged as a promising alternative to Transformers for sequence modeling in language tasks, addressing the quadratic time complexity of Transformers and their limitations in handling long contexts.
SSMs originate from classical control theory and are particularly inspired by continuous-time linear dynamical systems.
By discretizing the continuous-time formulation, SSMs can be expressed as discrete-time models that update the hidden state $s_t$ via a linear recurrence:
\begin{equation}
    s_t = A s_{t-1} + B x_t,
\end{equation}
where $x_t$ is the input and $A$ and $B$ are the state transition matrices.
Recent SSMs aim to design the transition matrices $A$, $B$ and the recurrence formulation.
S4~\cite{s4_gu}, H3~\cite{h3_fu}, S5~\cite{s5_smith}, Mamba~\cite{mamba_gu}, and Longhorn~\cite{longhorn_liu} have introduced structured state transition matrices and parallel computation schemes to enhance efficiency and capabilities.
In particular, recent SSM architectures enable linear-time inference and demonstrate strong extrapolation capabilities over long-range contexts, while achieving comparable performance to Transformers in language modeling tasks.

As SSMs have evolved, their applications have expanded beyond language modeling to various other domains, including robotics.
For instance, S5 is applied to reinforcement learning by allowing hidden state resets within a trajectory~\cite{s4rl_lu}.
MAIL~\cite{mail_jia} proposes a novel imitation learning policy by leveraging Mamba.
Building on these extensions, we explore using Longhorn, a recent state-of-the-art model, to perform in-context imitation learning.

\begin{figure*}[!t]
  \centering
  {\includegraphics[width=0.95\textwidth]{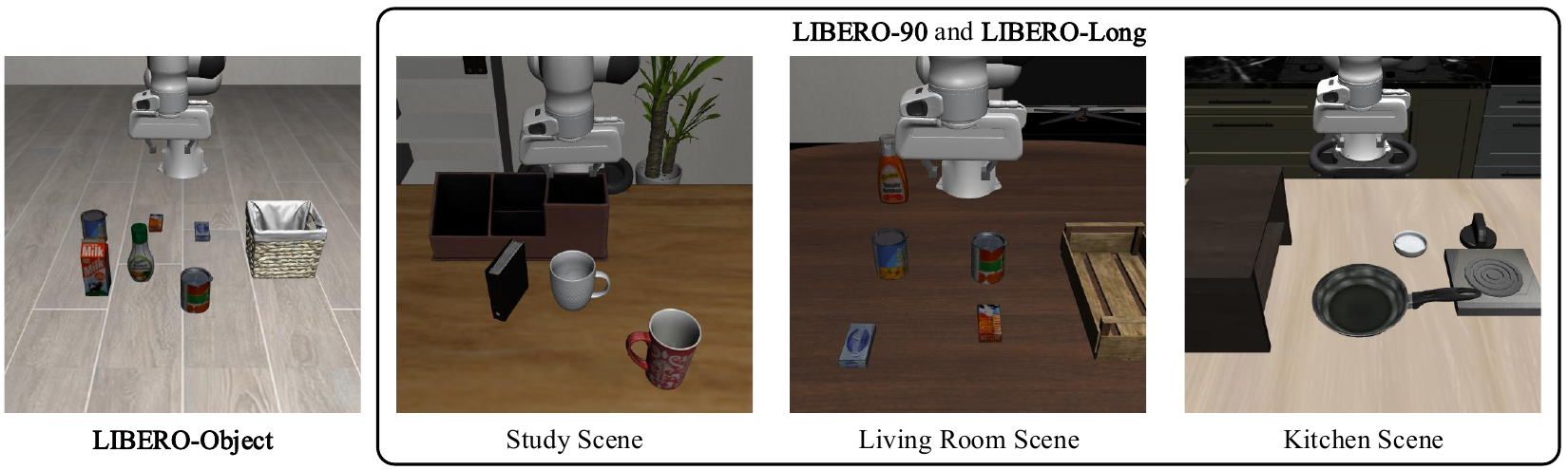}} 
    \caption{We evaluate \methodname{} on challenging manipulation tasks from the LIBERO benchmark. In LIBERO-90, the Study and Living Room suites each comprise 4 different scenes, while the Kitchen suite comprises 10 different scenes. Across these suites, the tasks cover diverse manipulation types, including pick-and-place tasks involving diverse objects and target locations (e.g., \textit{pick up the left/right bowl and place it in the tray} and \textit{pick up the book and place it in the back/front compartment of the caddy}), and drawer opening and closing tasks. We also evaluate on \textit{LIBERO-Long}, which contains long-horizon, multi-stage tasks (e.g., \textit{put both the alphabet soup and the cream cheese box in the basket}).}
    \label{fig:figure2}
\end{figure*}

\section{Method}

Our objective is to learn an ICIL policy $\pi_{\theta}$ that maximizes the success rate on unseen tasks when conditioned on few-shot demonstrations.
In this section, we describe how \methodname{} implements $\pi_{\theta}$ with Longhorn~\cite{longhorn_liu} and how it processes trajectory prompts during training and inference.

\subsection{Architecture}
\methodname{} first processes the observations with multimodal encoders. The per-step encoded observation embeddings are then passed through the Longhorn state-space block to generate actions.

\subsubsection{Input Encoding}
\methodname{} encodes multimodal observations at each time step of a demonstration.
Visual data are processed by convolutional neural networks (CNNs), and proprioceptive data are embedded using multi-layer perceptrons (MLPs).
The per-step features from each modality are concatenated and projected through an MLP to produce an observation embedding.
To prevent the model from trivially copying the actions in the prompt, we exclude actions from the input representation.
We further exclude any task language instructions to force the model to attend to the in-context demonstrations.

\subsubsection{Longhorn state-space block}
The sequence of observation embeddings $\{x_t\}_{t=1}^T$ is fed into Longhorn, which recurrently updates a memory state matrix $s_t \in \mathbb{R}^{d \times m}$.
At each time step, the input is interpreted as a key–value pair $(k_t, x_t)$, with $x_t \in \mathbb{R}^d$ and $k_t \in \mathbb{R}^m$ obtained via a linear projection of $x_t$, analogous to how Transformers use keys in the attention mechanism.
Longhorn then performs a recurrent update:
\begin{equation}
    s_t = A_t \odot s_{t-1} + B_t,
    \label{eq_update}
\end{equation}
where $\odot$ denotes the element-wise product, and $A_t, B_t : \mathbb{R}^d \rightarrow \mathbb{R}^{d \times m}$ are functions of $x_t$, defined as
\begin{subequations}\label{eq:ABeps}
\begin{align}
A_t &= (\mathbf{1}_{d\times m} - \varepsilon_t \otimes k_t^{\odot 2}), \label{eq:At}\\
B_t &= (\varepsilon_t \otimes \mathbf{1}_m) \otimes k_t, \label{eq:Bt}\\
\varepsilon_{t,i} &= \frac{\beta_{t,i}}{1 + \beta_{t,i}\, k_t^\top k_t}. \label{eq:Epsi}
\end{align}
\end{subequations}

where $\otimes$ denotes the outer product and $\beta_t \in \mathbb{R}^d$ is a weighting vector. 

From the updated state, we compute a context vector:
\begin{equation}
r_t = s_t q_t \in \mathbb{R}^d,
\label{eq:eq4}
\end{equation}
where $q_t \in \mathbb{R}^m$ is a query vector derived from a linear projection of $x_t$.

Finally, this context vector is passed through an output head to produce the corresponding action $a_t$.

\subsubsection{Longhorn for In-Context Imitation Learning}
From an online-learning perspective, the recurrent form in~\eqref{eq_update} can be derived as the solution to the following online convex programming objective~\cite{ocp_zinkevich}:
\begin{equation}
    s_t = \arg\min_{s \in \mathbb{R}^{d \times m}} \left\{ \|s - s_{t-1}\|_{\mathrm{F}}^2 + \|s k_t - x_t\|_{\mathrm{diag}(\beta_t)}^2 \right\},
    \label{eq2}
\end{equation}
where $\|\cdot\|_\mathrm{F}$ is the Frobenius norm and $\beta_t \in \mathbb{R}^d$ is a weighting vector.
This objective balances two competing goals inherent in online learning: the first term encourages the updated state $s_t$ to remain close to the previous state $s_{t-1}$, while the second term enforces that the current state $s_t$ accurately reflects the new input, allowing the model to incorporate newly observed information.
The weighting vector $\beta_t$ modulates this trade-off by controlling the relative importance of the current observation embedding; it is obtained by applying a sigmoid activation to a linear projection of the input $x_t$.

In this online regression view, the use of $\beta_t$ in equation~\eqref{eq2} naturally mitigates forgetting while integrating new information, thereby enabling efficient in-context learning with long contexts.
To encourage the model to attend to the demonstration prompts in ICIL, we scale $\beta$ at test time as $\beta'_t=\gamma\,\beta_t$, where $\gamma\in(0,1]$.

\begin{figure*}[!t]
  \centering
  {\includegraphics[width=0.95\textwidth]{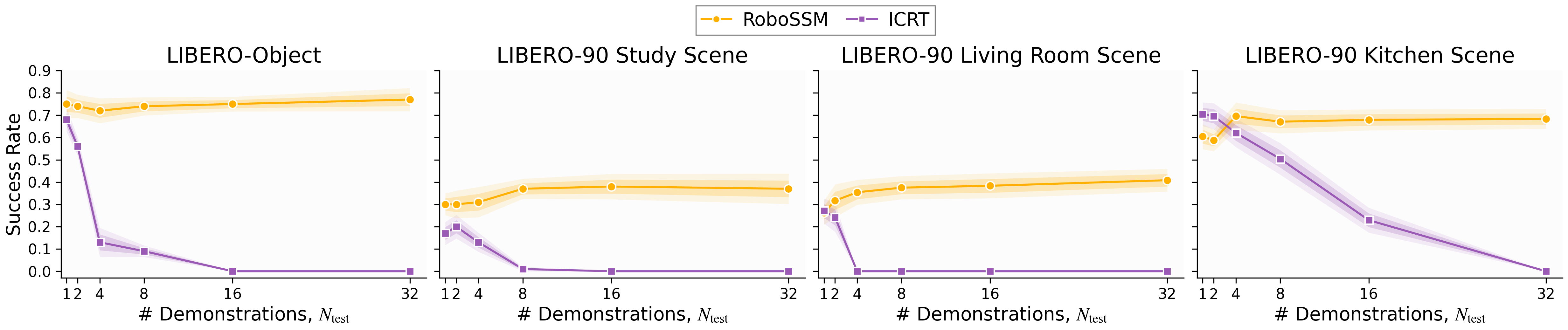}} 
    \caption{Comparison of \methodname{} and ICRT across test-time demonstrations ($N_\mathrm{test}$), with both models trained with $N_{\mathrm{train}}{=}2$. ICRT's performance drops sharply once $N_{\mathrm{test}}{>}N_{\mathrm{train}}$.}
    \label{fig:figure3}
\end{figure*}

\begin{figure*}[!t]
  \centering
  {\includegraphics[width=0.95\textwidth]{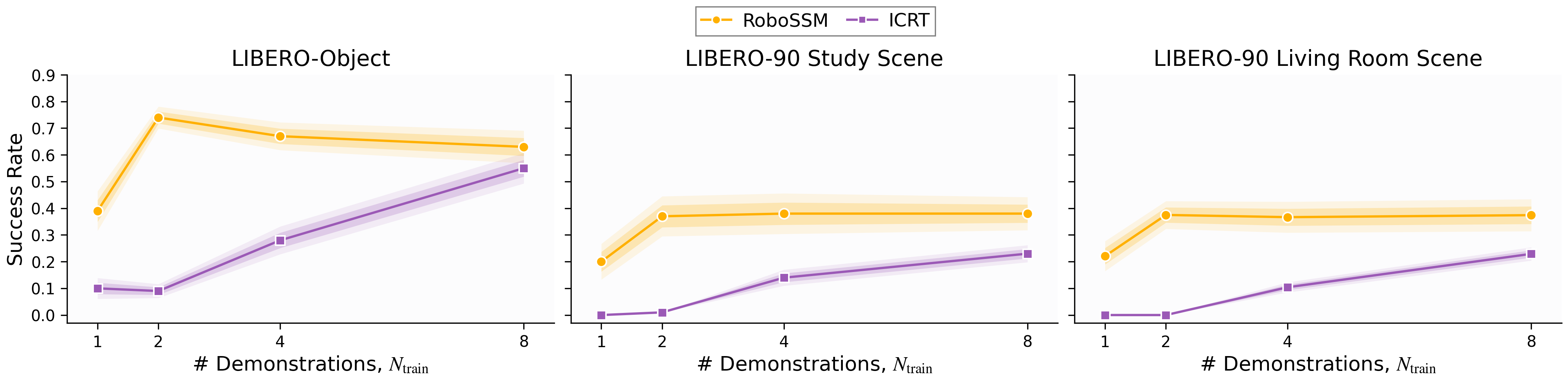}} 
  \caption{Results with a fixed test-time prompt $N_\mathrm{test} = 8$ across models trained with different numbers of demonstrations ($N_\mathrm{train}$).}
  \label{fig:figure4}
\end{figure*}

\begin{figure*}[!t]
  \centering
  {\includegraphics[width=0.95\textwidth]{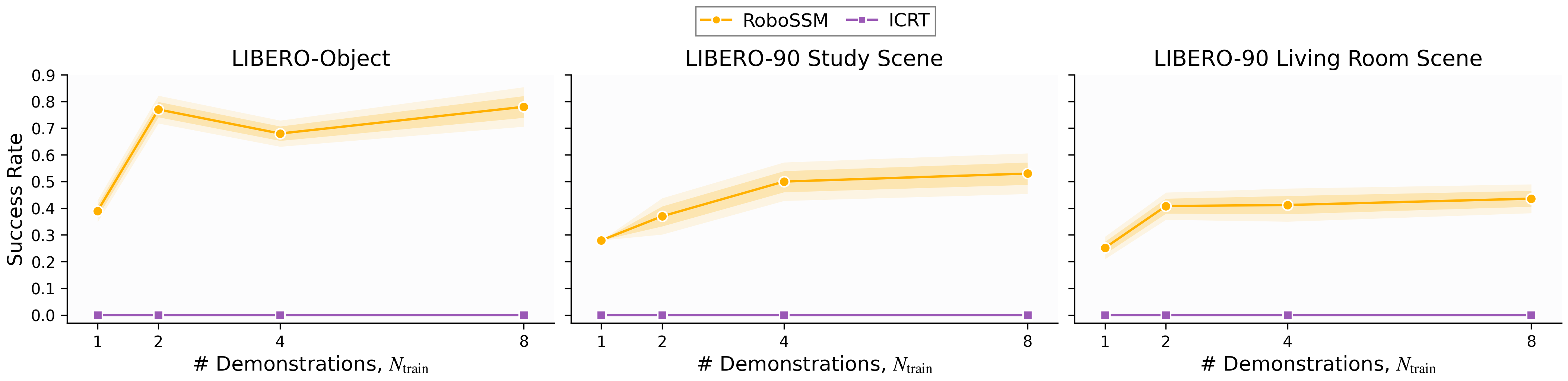}} 
  \caption{Results with a fixed test-time prompt $N_\mathrm{test} = 32$ across models trained with different numbers of demonstrations ($N_\mathrm{train}$). ICRT yields zero success across all $N_\mathrm{train}$.}
  \label{fig:figure5}
\end{figure*}

\subsection{In-Context Imitation Learning}

Following the standard ICIL formulation, \methodname{} conditions on a context of demonstration trajectories during both training and inference.
At test time, the policy adapts to new tasks based solely on the provided demonstrations, without any parameter updates.

Each input to the policy consists of a prompt and a query trajectory. 
The prompt $\mathcal{P}$ contains $N$ trajectories that provide task context:
\begin{equation}
    \mathcal{P} = \left[ \tau_1, \tau_2, \dots, \tau_{N} \right],
\end{equation}
where each demonstration trajectory $\tau_i$ is a sequence of $T_i$ observation embeddings:
\begin{equation}
    \tau_i = \left\{ o_1^{(i)}, o_2^{(i)}, \dots, o_{T_i}^{(i)} \right\},
\end{equation}
and $o_t^{(i)}$ denotes the $t$-th observation embedding from the $i$-th demonstration.  

Figure~\ref{fig:figure1} illustrates the training and inference procedure of the policy $\pi_{\theta}$. 
At each time step $t$, given a prompt $\mathcal{P}$ and the sequence of query observation embeddings $o_{1:t}^{q}$, the policy predicts the next action for the query trajectory as:
\begin{equation}
    a_t = \pi_{\theta}\!\left(\mathcal{P}; o_1^{q}, \ldots, o_t^{q}\right), \quad t = 1, \ldots, T_q.
\end{equation}

Both the prompt and the query trajectory are sampled from demonstrations of the same task. During training, the output actions of both the prompt and the query are supervised using the ground-truth actions.
We adopt a multi-task learning approach following ICRT~\cite{icrt_fu}, enabling the policy to infer the task intent from the prompt and to generalize to unseen tasks.

At inference time, the query trajectory is initialized with the first observation embedding $o_0^q$.
The policy then iteratively outputs the next action using the contextual information in $\mathcal{P}$ and applies the action to the environment, gradually building on the resulting observations until the task is complete.

\begin{figure*}[!t]
  \centering
  {\includegraphics[width=0.95\textwidth]{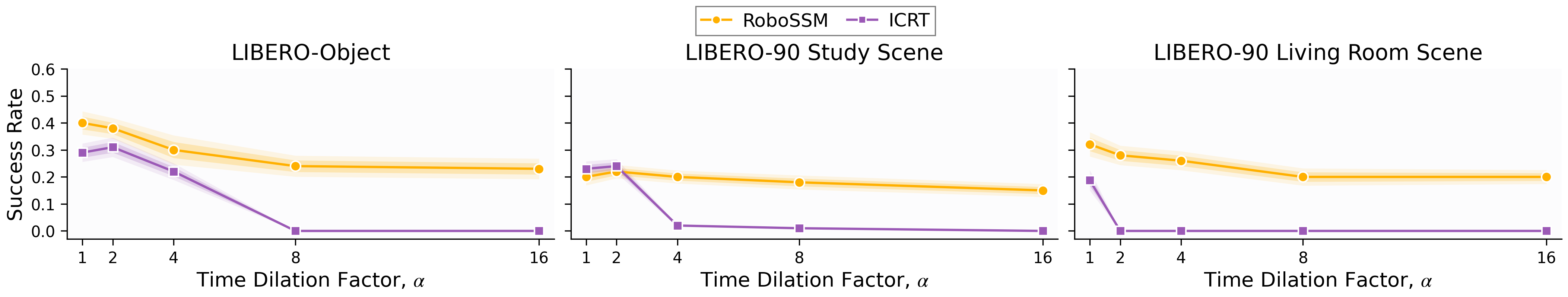}} 
  \caption{Comparison of \methodname{} and ICRT when evaluated on test-time prompts with temporally dilated demonstrations.}
  \label{fig:figure6}
\end{figure*}

\begin{figure*}[!t]
  \centering
  {\includegraphics[width=0.95\textwidth]{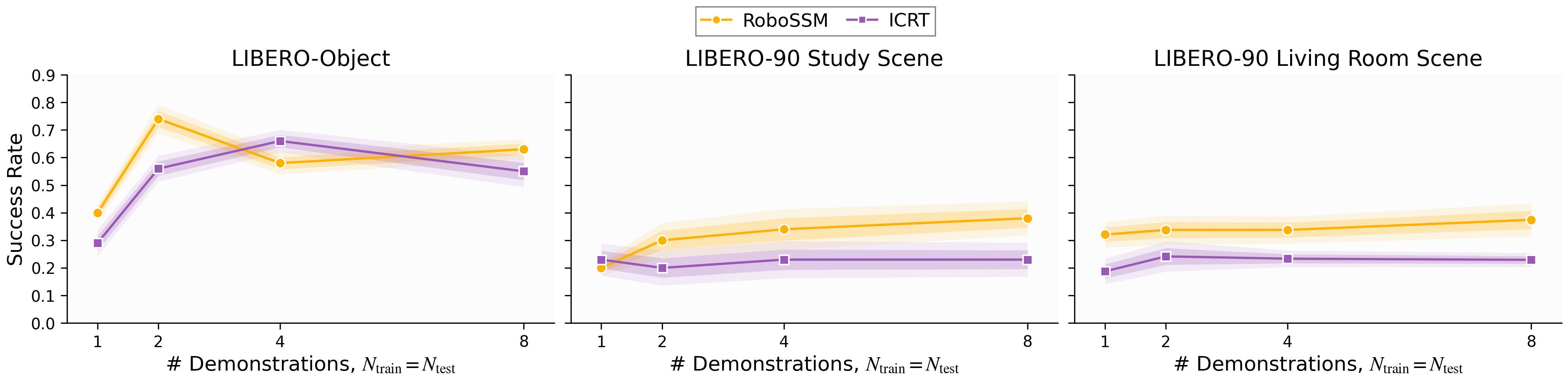}} 
  \caption{Results of \methodname{} and ICRT on in-context imitation learning, where the prompt consists of the same number of demonstrations during training and testing ($N_\mathrm{train} = N_\mathrm{test}$).}
  \label{fig:figure7}
\end{figure*}

\section{Empirical Results}

In this section, we evaluate whether \methodname{} can execute unseen tasks based on demonstration prompts, comparing it to previous state-of-the-art ICIL methods.
Section~\ref{sec:4.1} describes the experimental setup, including dataset construction.
We consider two regimes: out-of-distribution to assess length generalization with $|\mathcal{P}_{\mathrm{test}}|> |\mathcal{P}_{\mathrm{train}}|$ (Sec. \ref{sec:4.2}), and in-distribution where $|\mathcal{P}_{\mathrm{test}}|= |\mathcal{P}_{\mathrm{train}}|$ (Sec. \ref{sec:4.3}).
We then compare \methodname{} against a multi-task learning (MTL) policy without in-context learning on unseen tasks (Sec.~\ref{sec:4.4}) and investigate the effect of $\beta$-scaling in Longhorn for ICIL (Sec.~\ref{sec:4.5}).
Additionally, we provide latent-space visualizations of trajectories (Sec.~\ref{sec:4.6}) and an inference runtime analysis (Sec.~\ref{sec:4.7}).

\begin{table}[t]
\renewcommand{\arraystretch}{1.2}
\centering
\caption{Comparison of RoboSSM and ICRT on the \textit{LIBERO-Long living room} task suite, which consists of tasks of the form: ``Put both \textit{Object A} and \textit{Object B} in \textit{Object C}.'' Success rates are computed by assigning 0.5 for each correctly placed object and 1.0 for complete success.}
\begin{tabular}{lcc}
\hline
\multirow{2}{*}{Method} & \multicolumn{2}{c}{Success rate (\%)}                \\ \cline{2-3} 
                        & $N_{\text{test}} = 2$                     & $N_{\text{test}} = 4$                      \\ \hline
ICRT~\cite{icrt_fu}                   &0.0 $\pm$ 0.0                   & 0.0 $\pm$ 0.0                    \\
RoboSSM                 & \textbf{16.1 $\pm$ 0.6 } & \textbf{17.4 $\pm$ 0.9 }  \\ \hline
\end{tabular}
\label{table1}

\end{table}


We design our experiments to answer the following research questions:
\begin{itemize}
    \item \textbf{Q1:} Can \methodname{} extrapolate to prompts composed of demonstrations that are longer than those used in training?
    \item \textbf{Q2:} How many training demonstrations are required for \methodname{} to effectively infer with long prompts?
    \item \textbf{Q3:} Can \methodname{} achieve comparable performance to Transformer-based baselines when the test-time prompt length is equal to that used in training?
    \item \textbf{Q4:} Can \methodname{} achieve superior performance on unseen tasks compared to multi-task learning?
\end{itemize}

\subsection{Experimental Setup}
\label{sec:4.1}

\subsubsection{Datasets}
We conduct experiments on the \textbf{LIBERO} benchmark~\cite{libero_liu}, a challenging benchmark for visuomotor robot manipulation.  
LIBERO consists of five task suites: \textit{LIBERO-Object}, \textit{LIBERO-Goal}, 
\textit{LIBERO-Spatial}, \textit{LIBERO-Long}, and \textit{LIBERO-90}.  
\textit{LIBERO-90} consists of 90 tasks, while each of the other suites contains 10 tasks. Each task contains 50 demonstration trajectories.

In our experiments, we use the full \textit{LIBERO-Object} suite and divide \textit{LIBERO-90} into three task suites based on scene type, including \textit{kitchen}, \textit{living room}, and \textit{study} scenes, as shown in Figure~\ref{fig:figure2}.
The \textit{kitchen} suite contains drawer opening and closing tasks, and the other suites contain pick-and-place tasks.

For all experiments, the test set $\mathcal{D}_\text{test}$ contains tasks that are completely disjoint from the training set $\mathcal{D}_\text{train}$, ensuring that models are evaluated on entirely unseen tasks.
For training, we used \( |\mathcal{D}_{\text{train}}| = 8 \) for the \textit{LIBERO-Object} suite and for the \textit{kitchen} and \textit{study} suites, and \( |\mathcal{D}_{\text{train}}| = 14 \) for the \textit{living room} suite.
We set \( |\mathcal{D}_{\text{test}}| = 2 \) for all task suites.
For LIBERO-Long, we evaluated two long-horizon tasks using a model trained on the corresponding LIBERO-90 suite.

\subsubsection{Baselines}
We compare \textbf{\methodname{}} with \textbf{ICRT}~\cite{icrt_fu}, a Transformer-based in-context imitation learning method that employs LLaMA2-Base~\cite{llama2_touvron} as its backbone.
To ensure a fair comparison, we configure both backbones to have a similar number of parameters.
Both baselines use 4 blocks from their respective backbones. 
We use LLaMA2-Base with 6 attention heads per layer and a hidden size of 512. 
For Longhorn, the input is projected to a value dimension \(d=512\) and the keys and queries are projected to \(m=16\).
We also implement multi-task learning (MTL) without in-context learning for each backbone, where \textbf{MTL-TF} uses LLaMA2-Base and \textbf{MTL-SSM} uses Longhorn.
Unlike \methodname{} and ICRT, these MTL baselines take language instructions as input to specify the task.

\begin{table*}[t]
\centering
\small
\renewcommand{\arraystretch}{1.1}
\setlength{\tabcolsep}{6pt}

\caption{Comparison of \methodname{} and ICRT with multi-task learning (MTL) policies using their respective backbones. \textit{w/ lang} denotes that language instructions are included in the input, and \textit{w/o lang} denotes that language instructions are excluded. Both ICIL frameworks consistently outperform the MTL baselines, demonstrating the capabilities of in-context imitation learning methods.
}

\begin{tabular}{l *{4}{c}}
\hline
& & \multicolumn{3}{c}{LIBERO-90} \\
\cline{3-5}
Method & LIBERO-Object & Study Scene & Living Room Scene & Kitchen Scene \\
\hline
MTL-TF          & 24.6 $\pm$ 1.7 & 14.6 $\pm$ 6.4 & 2.5 $\pm$ 3.5 & 0.0 $\pm$ 0.0 \\
MTL-SSM          & 20.4 $\pm$ 3.3 & 15.0 $\pm$ 4.3 & 0.4 $\pm$ 0.9 & 0.0 $\pm$ 0.0 \\
\hline
ICRT~\cite{icrt_fu} & & & & \\
\quad w/o lang & \textbf{66.3 $\pm$ 3.8} & 22.5 $\pm$ 6.1 & 23.3 $\pm$ 2.8 & 0.0 $\pm$ 0.0 \\
\quad w/ lang                         & 48.3 $\pm$ 5.9 & 21.2 $\pm$ 3.8 & 27.1 $\pm$ 3.7 & 0.0 $\pm$ 0.0 \\
RoboSSM & & & & \\
\quad w/o lang             & 57.9 $\pm$ 3.5 & 34.2 $\pm$ 3.7 & \textbf{33.8 $\pm$ 4.5} & \textbf{3.8 $\pm$ 1.3} \\
\quad w/ lang       & 45.4 $\pm$ 2.3 & \textbf{35.4 $\pm$ 4.7} & 29.2 $\pm$ 5.3 & 0.0 $\pm$ 0.0 \\
\hline
\end{tabular}

\label{table2}
\end{table*}


\subsubsection{Implementation Details}
We use a single NVIDIA A100 GPU for all models and task suites. 
During training, we use the AdamW~\cite{adamw_loshchilov} optimizer with weight decay 1e-4, $\beta_1=0.9$, and $\beta_2=0.999$. 
The learning rate follows a cosine decay schedule from 1e-4 to 1e-5. 
We use front-view and hand-view RGB images for visual observations, and the robot’s joint angles and gripper state for proprioception.
For visual observations, data augmentation includes color jitter with brightness, contrast, and saturation factors of 0.3, and a random masking scheme that applies up to 8 square masks of size $16\times16$. 
Models are trained for 200 epochs with a batch size of 4. 
During evaluation, we execute 20 rollouts per task, generate trajectories of up to 200 time steps, and report the average success rate over all tasks in the suite across 6 random seeds.




\subsection{Prompt Length Generalization}
\label{sec:4.2}
We investigate long-range in-context imitation learning with \methodname{}, focusing on prompt-length extrapolation to out-of-distribution contexts.
In this section, we consider three approaches to making the test prompt substantially longer than the training prompt ($|\mathcal{P}_{\mathrm{test}}|> |\mathcal{P}_{\mathrm{train}}|$): (1) increasing the number of demonstrations, (2) evaluating on long-horizon tasks, and (3) applying temporal dilation to demonstrations.

\subsubsection{Number of demonstrations}
To measure how performance changes when the number of test-time demonstrations differs from that of training, we train models with a small number of demonstrations $N_\mathrm{train} = 2$.
We then evaluate them on test prompts with $N_\mathrm{test}\in\{1, 2, 4, 8, 16, 32\}$. 
As shown in Figure~\ref{fig:figure3}, \methodname{} improves and then maintains its success rates as $N_{\mathrm{test}}$ increases beyond $N_{\mathrm{train}}$.
On LIBERO-Object, \methodname{} achieves its best performance at $N_{\mathrm{test}} = 32$, which is 16 times longer than the training prompt length $|\mathcal{P}_{\mathrm{train}}|$, despite never having observed such long prompts during training.
This result answers \textbf{Q1} affirmatively, showing that \textbf{\methodname{} extrapolates effectively to prompts far beyond the training horizon, thereby achieves high success rates}.
In contrast, ICRT degrades sharply once $N_{\mathrm{test}} > N_{\mathrm{train}}$ and collapses on the long prompts.
This result suggests that ICRT fails to generalize to longer prompts, performing reliably only when $N_{\mathrm{test}}$ is equal to or shorter than $N_{\mathrm{train}}$.

We further investigate how many training demonstrations are required for long-range in-context learning.
We train models with $N_{\mathrm{train}}\in\{1,2,4,8\}$ and evaluate them with fixed test-time prompts $N_{\mathrm{test}}\in\{8,32\}$.
According to Figure~\ref{fig:figure4},
\methodname{} maintains high success rates even when trained with few task demonstrations, whereas ICRT collapses to 0\% success rate for all $N_{\mathrm{train}}$.
Figure~\ref{fig:figure5} shows that ICRT is comparable to \methodname{} only when trained with demonstrations matching the test-time prompt length.
These results show that \textbf{\methodname{} achieves strong long-range in-context learning even with limited training demonstrations, highlighting its training efficiency} and thereby addressing \textbf{Q2}.


\subsubsection{Long-horizon task adaptation}
Demonstration prompts constructed from long-horizon tasks are inherently longer than prompts constructed from short-horizon tasks. 
Consequently, generalization to long-horizon tasks via ICIL beyond the short-horizon training distribution requires context-length extrapolation.
We evaluate this capability on long-horizon tasks with $N_{\mathrm{test}} \in \{2, 4\}$, using models trained on short-horizon tasks with $N_{\mathrm{train}} = 2$.

Table~\ref{table1} suggests that RoboSSM can perform challenging long-horizon tasks by handling long demonstration prompts despite being trained only on short-horizon task suites (\textbf{Q1}), whereas ICRT fails in all test rollouts.
These results demonstrate that \textbf{\methodname{} enables generalization to long-horizon tasks that Transformer-based ICIL fails to solve.}
        
\subsubsection{Time dilation}
In real-world scenarios, robot demonstrations may vary in execution speed due to factors such as operator latency, hardware variability, or differing task conditions.
To simulate such temporal variability, we evaluate whether the model can generalize to time-dilated demonstrations at test time.
We create temporally stretched demonstrations by repeating each observation embedding in the original trajectory $\alpha$ times, resulting in a new trajectory of length $\alpha \cdot T$, where $T$ is the original trajectory length and $\alpha \in \{1, 2, 4, 8, 16\}$ is the dilation factor.
Although models are trained with the original prompt length ($\alpha = 1$), we evaluate their robustness under extended test-time prompts, where $|\mathcal{P}_{\mathrm{test}}| = \alpha \cdot |\mathcal{P}_{\mathrm{train}}|$, with $\alpha$ up to 16.

Figure~\ref{fig:figure6} indicates that \methodname{} sustains competitive success rates across increasing dilation factors $\alpha$, highlighting robustness to temporal stretching and long-context extrapolation (\textbf{Q1}).
By contrast, ICRT exhibits a consistent performance decay with $\alpha$, culminating in failure on the longest prompts.
These results demonstrate that \methodname{} can be adopted in real-world applications of ICIL 
These results suggest that \textbf{\methodname{} supports the real-world deployment of ICIL by handling prompt length increases caused by temporal variation in demonstrations}.

\begin{figure}[t]
  \centering
  {\includegraphics[width=0.85\columnwidth]{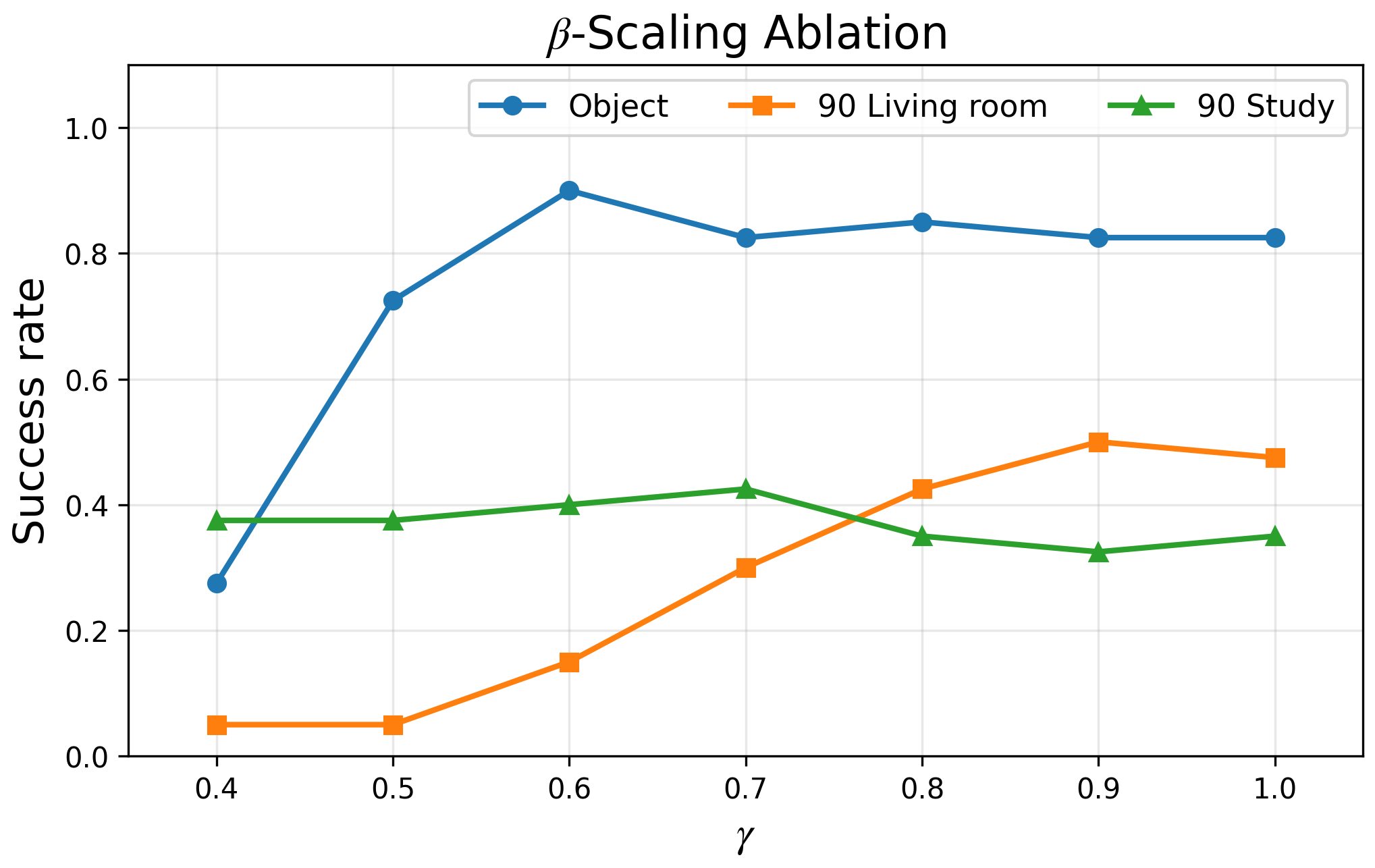}} 
  \caption{Effectiveness of test-time $\beta$-scaling across task suites.}
  \label{fig:figure8}
\end{figure}

\begin{figure}[t]
  \centering
  \subfloat[ICRT\label{fig:figure9_a}]{%
    \includegraphics[width=0.8\columnwidth]{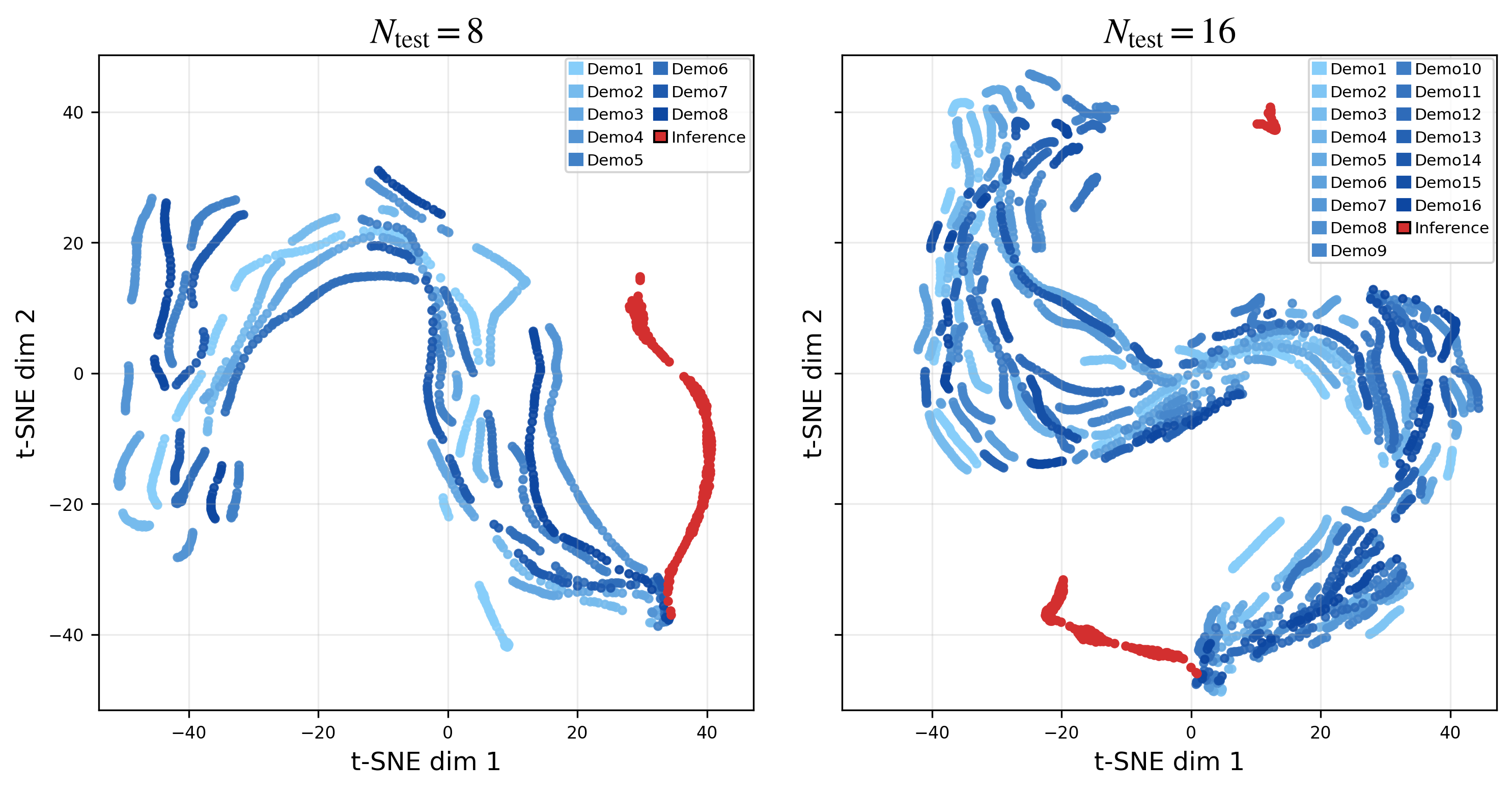}
  }\par\vspace{-1pt}
  \subfloat[\methodname{}\label{fig:figure9_b}]{%
    \includegraphics[width=0.8\columnwidth]{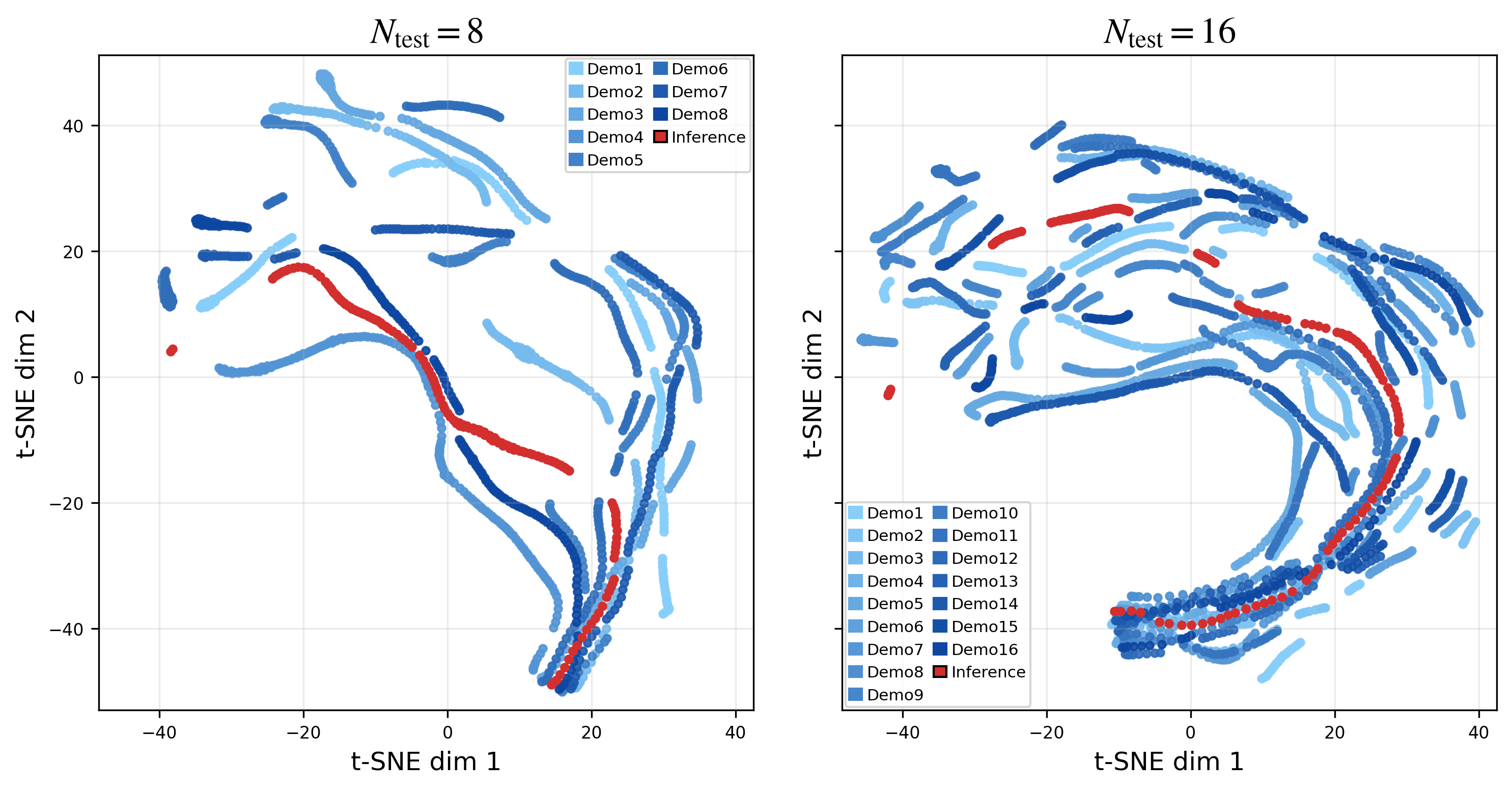}
  }
  \caption{Visualization of latent representations with \(N_{\mathrm{test}}=8\) (left) and \(N_{\mathrm{test}}=16\) (right). Blue corresponds to prompt demonstrations and red to the predicted trajectory.}
  \label{fig:figure9}
  \vspace{-20pt}
\end{figure}

\subsection{In-distribution ICIL}
\label{sec:4.3}

We evaluate \methodname{} and ICRT on in-distribution tasks, where $N_\mathrm{train} = N_\mathrm{test} \in \{1,2,4,8\}$, covering varying numbers of demonstrations.
This evaluation addresses \textbf{Q3} when both models operate under distributional conditions similar to those in training.
Given that Longhorn exhibits performance parity with Transformers in language modeling tasks, we expect it to demonstrate comparable performance to Transformers in this setting.
Nevertheless, as shown in Figure~\ref{fig:figure7}, \methodname{} consistently achieves higher success rates than ICRT across most scenarios, particularly in the LIBERO-90 Study and Living Room scenes.
This performance is likely due to the Longhorn architecture, whose formulation as an online regression problem enhances its in-context learning capability.

\subsection{Comparison to Multi-Task Learning}
\label{sec:4.4}
We compare \methodname{} against multi-task learning baselines, MTL-TF and MTL-SSM.
During training, we set $N_\mathrm{train}=4$, and for evaluation, $N_\mathrm{test}=0$ for MTL baselines, while $N_\mathrm{test}=4$ for \methodname{} and ICRT.
To enable a fair comparison with language-conditioned MTL, we additionally train and evaluate \methodname{} and ICRT with language instructions.
In response to \textbf{Q4}, Table~\ref{table2} shows that \methodname{} and ICRT reliably surpass the MTL baselines across their respective backbone architectures.
However, the inclusion of language does not lead to improved performance on unseen tasks, as the language instructions serve to identify tasks~\cite{libero_liu}.
These results demonstrate that \textbf{\methodname{} effectively handles unseen tasks without any parameter updates and that \methodname{} achieves stronger performance compared to prior methods}.

\begin{figure}[t]
  \centering
  {\includegraphics[width=0.8\columnwidth]{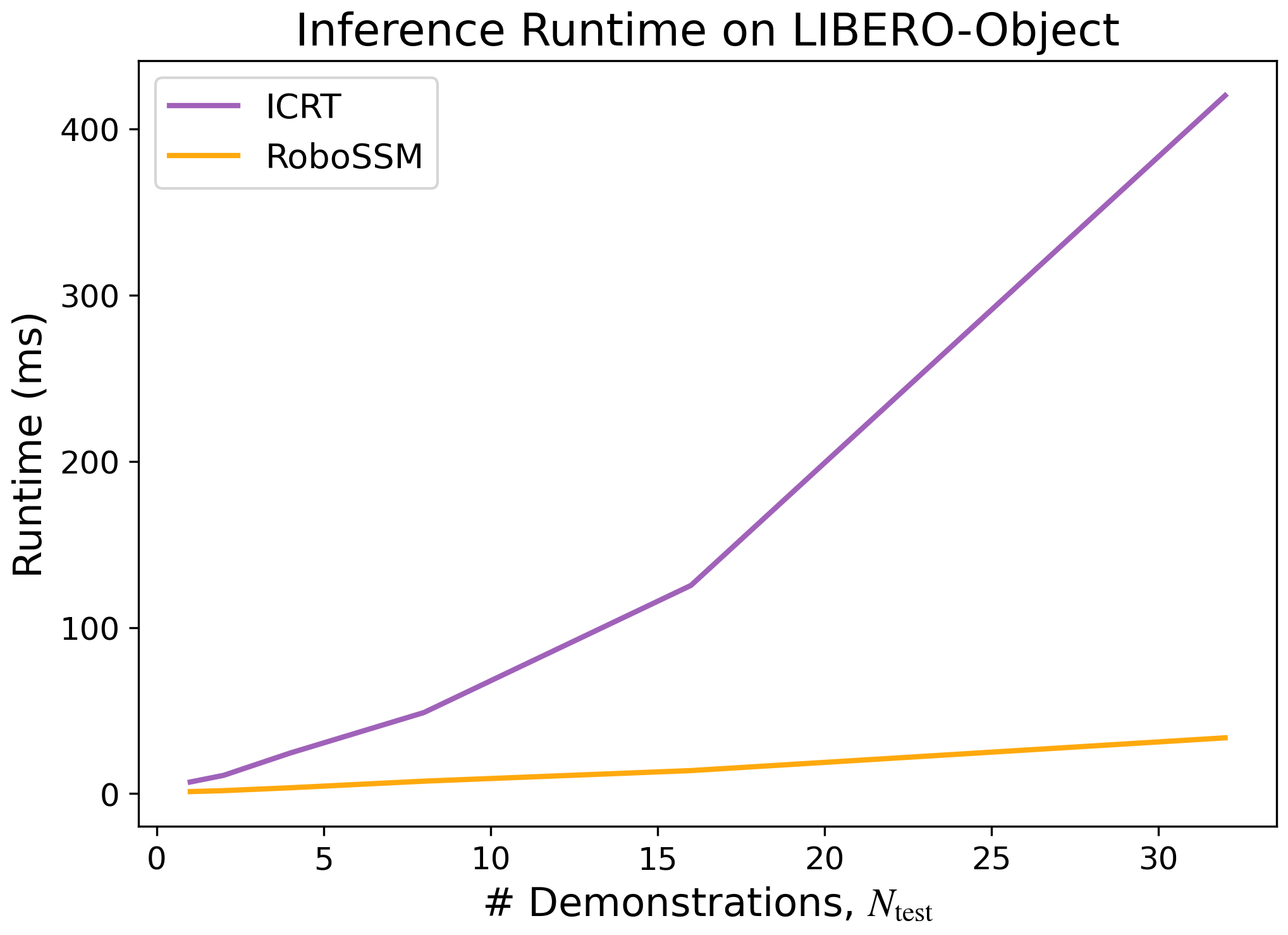}} 
  \caption{Inference runtime on LIBERO-Object for ICRT and \methodname{}.}
  \label{fig:figure10}
  \vspace{-20pt}
\end{figure}

\subsection{$\beta$-Scaling of Longhorn for ICIL}
\label{sec:4.5}
We replace the weight vector $\beta_t$ in equation~\eqref{eq2} with a test-time rescaled weight vector $\beta'*t=\gamma,\beta_t$ to investigate the effect of regulating the trade-off between retaining prior state information and incorporating new inputs in \methodname{}.
Figure~\ref{fig:figure8} reports results for $N_{\mathrm{test}}=8$ and $N_{\mathrm{train}}=2$ with the scaling factor $\gamma\in[0.4,1.0]$.
The results show that scaling the weight vector $\beta_t$ through $\gamma$ affects performance on unseen tasks, with the best scaling factor varying across task suites.

\subsection{Latent Space Analysis}
\label{sec:4.6}
Figure~\ref{fig:figure9} displays visualizations of per-timestep latent representations using two-dimensional t-SNE~\cite{gradients_metz}, covering both the prompt demonstrations and the subsequent predicted trajectory. 
For ICRT, we plot the MLP output from the final LLaMA2-Base block.
For \methodname{}, we plot the query-projected state from the final Longhorn block, as defined in equation~\eqref{eq:eq4}.
\methodname{} generates trajectories that stay close to the prompt demonstration clusters, indicating stable long-context extrapolation.
In contrast, ICRT produces trajectories that drift far outside the prompt manifold, with the deviation becoming more pronounced as the number of prompt demonstrations increases.

\subsection{Runtime Analysis}
\label{sec:4.7}
Figure~\ref{fig:figure10} reports the inference runtime on LIBERO-Object. Each run uses a prompt constructed from $N_{\mathrm{test}}$ demonstrations and executes $T_q=200$ action-update steps. \methodname{} achieves lower runtime, whereas ICRT’s runtime increases rapidly as the number of demonstrations grows. \methodname{} scales nearly linearly with the prompt length $L$, yielding total work $\mathcal{O}(L + T_q)$. In contrast, ICRT, with a LLaMA-2 backbone and a KV cache, incurs $\mathcal{O}(L^{2})$ prefill over the prompt and $\mathcal{O}(T_q)$ decoding for next-action prediction. Its runtime therefore increases markedly more rapidly as $L$ grows, widening the runtime gap in favor of \methodname{} at longer prompt lengths. Overall, \textbf{\methodname{} maintains efficient inference with long-context demonstrations}.

\section{Conclusion}
In this work, we introduce \methodname{}, a scalable in-context imitation-learning framework built on state-space models (SSMs).
\methodname{} executes unseen tasks via few-shot prompts with strong prompt-length extrapolation and linear-time inference.
Across the LIBERO benchmarks, \methodname{} can process prompts up to 16 times longer than those seen during training, which enables it to (1) outperform Transformer-based ICIL methods under increased numbers of demonstrations in the prompt and (2) generalize to long-horizon tasks.
Building on these results, \methodname{} can support continual adaptation for lifelong learning by simply being fed demonstration prompts for new tasks, without any task-specific parameter updates.

While \methodname{} highlights the potential of SSMs as a promising backbone for long-context ICIL, comprehensively addressing novel tasks will require broader and more diverse training corpora. 
Future work may explore scaling datasets in both size and task diversity to enable more effective generalization to novel tasks. 
In addition, while validation of \methodname{} on real robots across diverse real-world tasks is not essential to the core contribution of this work, it could further demonstrate its generalization capability.

\bibliographystyle{IEEEtran}
\bibliography{IEEEabrv,refs}

\addtolength{\textheight}{-12cm}   

\end{document}